%% file: main.tex
\documentclass[runningheads]{llncs}

\usepackage{eccv}

\usepackage{eccvabbrv}

\usepackage{graphicx}
\usepackage{booktabs}
\usepackage{multirow}
\usepackage{todonotes}

\usepackage[accsupp]{axessibility}  %

\usepackage{hyperref}

\usepackage{orcidlink}

\begin{document}

\title{KAN You See It? \\
KANs and Sentinel for Effective and Explainable Crop Field Segmentation}

\titlerunning{KAN You See It?}

\author{Daniele Rege Cambrin\inst{1}\orcidlink{0000-0002-5067-2118} 
\and
Eleonora Poeta\inst{1}\orcidlink{0009-0001-7289-0036} %
\and
Eliana Pastor\inst{1}\orcidlink{0000-0002-3664-4137}
\and
Tania Cerquitelli\inst{1}\orcidlink{0000-0002-9039-6226}
\and
Elena Baralis\inst{1}\orcidlink{0000-0001-9231-467X}
\and
Paolo Garza\inst{1}\orcidlink{0000-0002-1263-7522}
}

\authorrunning{D. Rege Cambrin et al.}
\institute{Politecnico di Torino, Turin, Italy\\
\email{\{daniele.regecambrin,eleonora.poeta,eliana.pastor,\\tania.cerquitelli,elena.baralis,paolo.garza\}@polito.it}}

\maketitle

\begin{abstract}

    Segmentation of crop fields is essential for enhancing agricultural productivity, monitoring crop health, and promoting sustainable practices. Deep learning models adopted for this task must ensure accurate and reliable predictions to avoid economic losses and environmental impact.
    The newly proposed Kolmogorov-Arnold networks (KANs) offer promising advancements in the performance of neural networks.
    This paper analyzes the integration of KAN layers into the U-Net architecture (U-KAN) to segment crop fields using Sentinel-2 and Sentinel-1 satellite images and provides an analysis of the performance and explainability of these networks.  Our findings indicate a 2\% improvement in IoU compared to the traditional full-convolutional U-Net model in fewer GFLOPs. Furthermore, gradient-based explanation techniques show that U-KAN predictions are highly plausible and that the network has a very high ability to focus on the boundaries of cultivated areas rather than on the areas themselves. The per-channel relevance analysis also reveals that some channels are irrelevant to this task.
  \keywords{Remote Sensing \and Explainable AI \and Crop Field Segmentation}
\end{abstract}

\section{Introduction}
\label{sec:intro}

\begin{figure}[htb]
    \centering
    \subfloat[RGB Image]{\includegraphics[width=0.2\linewidth]{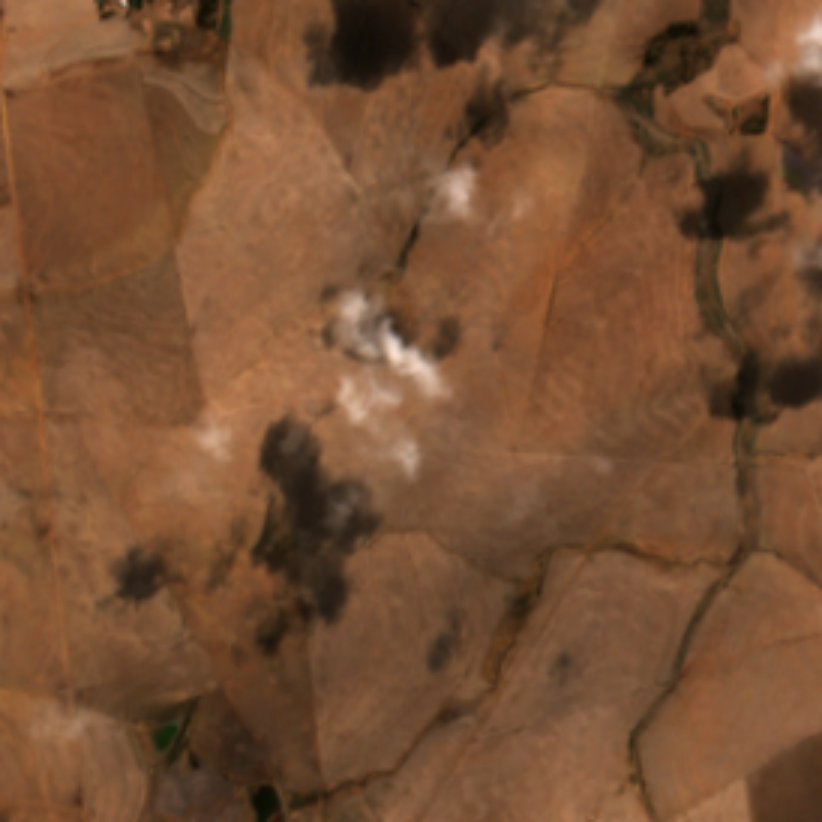}} \hfill
    \subfloat[Ground Truth]{\includegraphics[width=0.2\linewidth]{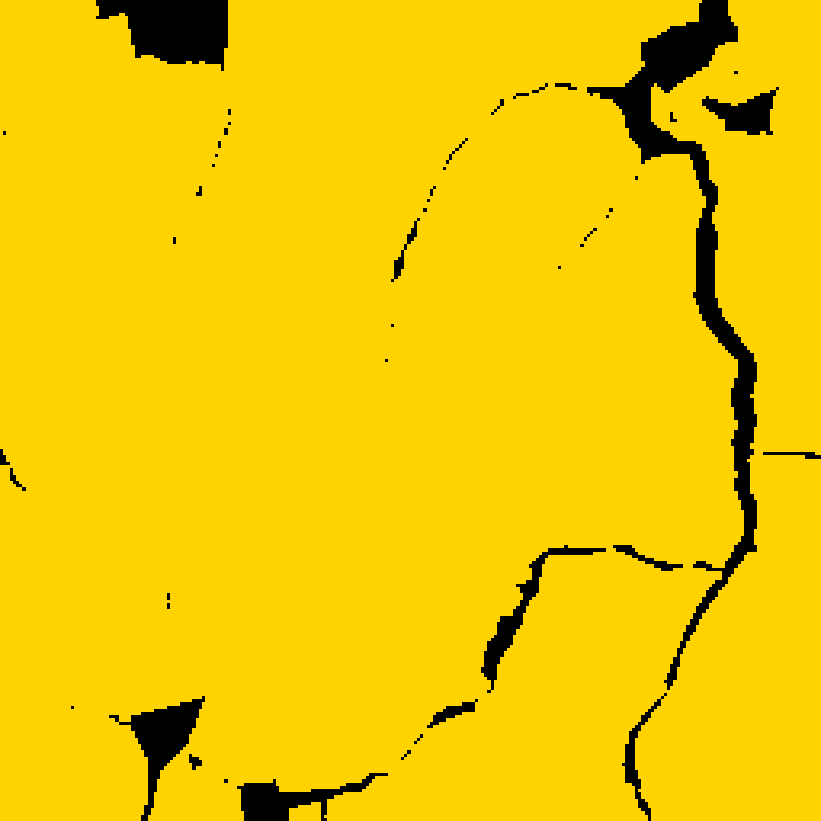}} \hfill
    \subfloat[U-Net]{\includegraphics[width=0.2\linewidth]{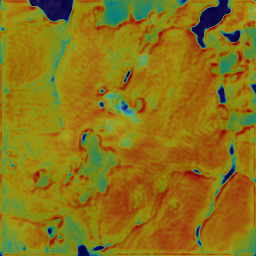}} \hfill
    \subfloat[U-KAN]{\includegraphics[width=0.2\linewidth]{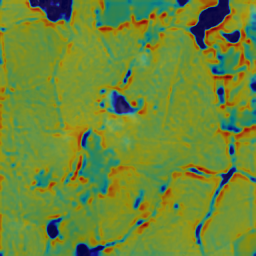}}
    \caption{Example image from the  South Africa Crop Type dataset~\cite{sa_crop_type_dataset}, Sentinel-2. (a) displays the image from Sentinel-2 in RGB, and (b) shows the corresponding ground truth, with crop field areas for segmentation highlighted in yellow. (c) and (d) present the saliency maps generated by U-Net and U-KAN, respectively, where red pixels indicate the areas of highest network focus.}
    \label{fig:saliency_comp}
\end{figure}

In recent years, remote sensing and deep neural networks have revolutionized how we approach agricultural management, environmental monitoring, and many earth-observation-related tasks. 
Their combination proved to be effective in a wide range of tasks, such as emergency management~\cite{disaster_management_review} and land cover~\cite{land_cover_review}. 
One task related to land cover is the segmentation of crop fields, which is crucial for optimizing agricultural productivity, assessing crop health, and planning sustainable farming practices~\cite{precision_farming_impact}. 

The accuracy and interpretability of the neural networks used in this process are fundamental to ensure reliable and actionable insights.
Accurate segmentation of crop fields enables precise calculations of area coverage, assessment of crop types, and monitoring of agronomic factors such as plant health and soil conditions~\cite{biotic_stress_review}. 
This information is critical for making informed decisions on irrigation, fertilization, and crop rotation, which are essential for enhancing yield and sustainability~\cite{crop_type_segmentation}. 
Furthermore, accuracy in semantic segmentation tasks directly influences economic planning and policy-making at various governmental and institutional levels. 
Providing accurate decisions is essential, yet model understandability and accessibility are also vital to allow practitioners to validate them and adhere to institutional regulations~\cite{unesco2021recommendation}. 
These factors are crucial as they significantly impact both economies and the environment.\cite{gevaert2022explainable}.
Deep learning models can achieve high accuracy, but they are often considered ``black boxes'' due to their intricate architectures composed of numerous layers and parameters that are difficult to interpret. 
This complexity poses significant challenges in understanding the decision-making processes underlying these models. 
In the context of remote sensing, the interpretability of such models is further complicated by the nature of the data, which includes various spectral bands, temporal sequences, and spatial resolutions. Additionally, factors such as noise, occlusions, and atmospheric effects can obscure the models' decision-making processes.

Consequently, explainability in deep learning for remote sensing is crucial, as it ensures that humans can understand the decisions and outputs of these models. 
Developing techniques to elucidate the logic behind the model outputs is essential for validating its results and establishing confidence in its practical applications.
A widely adopted approach to enable model explainability is to provide explanations for individual predictions of a model in a post-hoc fashion, allowing its interpretability while not affecting its accuracy. This solution finds application in the domain of earth observation, where explanations are presented as saliency maps (or heat maps), highlighting which parts of the satellite image influence the model prediction~\cite{kakogeorgiou2021evaluating, gevaert2022explainable} (see Figure~\ref{fig:saliency_comp} (c) and (d) as examples).
The recent introduction of Kolgomorov-Arnold Networks (KANs)~\cite{kan} posed a new paradigm for neural networks as an alternative to Multi-Layer Perceptrons (MLPs). 
Inspired by the Kolmogorov-Arnold representation theorem~\cite{kolmogorov1956representation,arnold1957functions}, KANs allow learning custom activations of the edge of the network. In this way, it is possible to analyze the contribution of individual components of the input data, thus providing a more transparent view of the network’s decision-making process. 
Considering their potential in improving vision tasks, KANs were recently integrated~\cite{ukan} into the U-Net architecture~\cite{unet}, which is a well-known segmentation architecture. 
The resulting network, denoted as U-KAN, was tested for medical image segmentation and demonstrated superior accuracy and efficiency.

In this paper, we are the first to explore the adoption of U-KAN for crop field segmentation, analyzing U-KAN and U-Net from the performance and explainability perspectives. 

We consider the following research questions (RQs):

\begin{description}
    \item  \textbf{RQ1.} How does U-KAN perform compared to U-Net for the task of crop field segmentation?
    \item \textbf{RQ2.} Which parts of the satellite image influence the models' predictions the most? Do U-Net and U-KAN prioritize different aspects of the image? 

\end{description}

To address the first research question, we evaluate U-KAN and U-Net on the South Africa Crop Type dataset~\cite{sa_crop_type_dataset} for crop field segmentation on Sentinel-2~\cite{sentinel2} and Sentinel-1~\cite{sentinel1} satellite images.
Our findings indicate that U-KANs are more accurate and efficient than U-Nets regarding Intersection-Over-Union (IoU) and Giga Floating Point Operations Per Second (GFLOPs), respectively.

For the second research question, we analyze the explainability of both models. 
We leverage post-hoc explainability techniques to identify which parts of an image influence the identification of the model.
We compare these importance scores, denoted as saliency maps, for U-Net and U-KAN to analyze differences in the identification behavior.
Our results reveal that U-Net and U-KAN indeed consider different aspects of the image for predictions. 
U-KANs generally focus on the edges of the crops, while U-Nets concentrate more on the internals, as exemplified in \Cref{fig:saliency_comp} (c) and (d). 
Moreover, we quantitatively evaluate the quality of the saliency maps, revealing that U-KANs are more faithful and plausible.

Our contributions can be summarized as follows:
\begin{itemize}
     \item We are the first to explore the application of U-KAN for crop field segmentation.
    \item We perform a comparative analysis of U-KAN and U-Net for the crop field segmentation on satellite images.
    \item We analyze the explainability of U-KAN and U-Net, utilizing post-hoc explainability techniques to generate and evaluate saliency maps.
    \item We reveal that U-KAN offers superior accuracy and efficiency, as well as more faithful and plausible saliency maps compared to U-Net.
\end{itemize}

The code to reproduce the experiment is available at \url{https://github.com/DarthReca/crop-field-segmentation-ukan}.

\section{Related Work}
\label{sec:related}
In this section, we provide an overview of the advancements in remote sensing for agriculture, the explainability of neural networks, and their intersection.

\subsection{Remote Sensing}
Remote sensing technologies have been extensively applied in agriculture to enhance crop monitoring, management, and productivity. Early studies focused on utilizing satellite imagery to assess crop health and estimate yields~\cite{spectral_inputs_crop}. With advancements in sensor technology and data processing techniques, the resolution and accuracy of remote sensing data have significantly improved, enabling more detailed analysis of agricultural landscapes~\cite{remote_sensing_agriculture_review}.
One application of remote sensing in agriculture is crop field segmentation, which involves identifying cultivated areas. The introduction of Convolutional Neural Networks (CNNs) and the U-Net architecture has further enhanced crop field segmentation~\cite{time_series_crop_classification,unet_crop_segmentation}.  Although recent advancements have proposed other architectures, it still remains one of the most effective baselines in remote sensing, thanks to its design~\cite{change_detection_reality}.
Integrating multispectral and hyperspectral imaging has also contributed to more accurate crop field segmentation. These images capture data across various wavelengths, providing richer information about crop characteristics~\cite{hyperspectral_vegetation}.

\subsection{Explainable AI}
Explainable artificial intelligence (XAI) is a branch of AI research dedicated to making machine learning models interpretable and understandable to humans
~\cite{arrieta2020explainable, adadi2018peeking, molnar2020interpretable}. 
In recent years, there has been a significant interest in applying XAI techniques to Earth observation tasks, driven by the necessity to interpret complex AI models applied in remote sensing~\cite{gevaert2022explainable, kakogeorgiou2021evaluating}.
Solutions in this domain follow a standard categorization of XAI approaches: interpretable by design and post-hoc explainability methods~\cite{molnar2020interpretable}. 
By-design approaches, such as~\cite{martinez2020crop, mateo2021learning, stomberg2021jungle, levering2020interpretable}, integrate interpretability intrinsically into the design of the model algorithm or its architecture. However, they often fail to explain individual model predictions, raising doubts about their actual ability to help humans understand the process~\cite{gevaert2022explainable}. Additionally, these approaches tend to be less accurate than black-box models. To address these limitations, many works focus on post-hoc explanations~\cite{maddy2021miidaps, huang2021better, hung2021integrating, kakogeorgiou2021evaluating, temenos2023interpretable}, which aim to explain trained black-box models while preserving their accuracy and enhancing transparency. 

Saliency maps are one of the most adopted post-hoc methods for visualizing which parts of an input image influence the model's prediction. 
Saliency maps (or heat maps) are overlayed pixel-based importance scores over the input image, highlighting how much each pixel contributes to the prediction. These maps have been widely adopted for semantic segmentation in tasks such as medical diagnosis~\cite{maddy2021miidaps, huang2021better, hung2021integrating, kakogeorgiou2021evaluating}. The urgency of understanding the decision process of the models in remote sensing has led to works investigating their application to segment satellite imagery and agricultural fields~\cite{kakogeorgiou2021evaluating}.

Among these works, Kakogeorgiou and Karantzalos~\cite{kakogeorgiou2021evaluating} proposed a systematic evaluation of saliency maps for multi-label deep learning classification tasks in remote sensing. Their work tested ten explainable AI methods for deriving saliency maps and proposed a systematic analysis from both qualitative and quantitative perspectives. Our approach aligns with this systematic evaluation; however, rather than comparing multiple XAI approaches to explain the same model, we systematically evaluate explanations by comparing two models using the same explainability technique to derive saliency maps. 
Their study identified Grad-CAM~\cite{selvaraju2016grad} as a reliable and interpretable method while being less computationally expensive~\cite{kakogeorgiou2021evaluating}. We leverage the result of this analysis, and we adopt Grad-CAM as the explainability method. 

In line with the demand for interpretability, the recently proposed KANs~\cite{kan} offer their own level of interpretability by allowing interactions with the network through pruning and by visualizing the learnable activation functions. 
The work~\cite{ukan} demonstrated the integration of KANs into the U-NET architecture~\cite{unet}, enhancing performance and efficiency for medical diagnosis tasks. Our study is the first to apply U-KANs for crop field segmentation. Furthermore, we compare the U-NET and U-KAN architectures from an explainability perspective, proposing a systematic evaluation of their explanations. To the best of our knowledge, we are also the first to evaluate U-KAN for the post-hoc explainability perspective, analyzing the individual explanations provided by Grad-CAM.

\section{Methodology}
\label{sec:method}
In this section, we formalize the problem by first detailing the crop field segmentation task, followed by the explainability part, and finally, we present the models.

\subsection{Problem statement}
This work addresses the crop field segmentation problem based on radiometric or hyperspectral images. The problem can be formulated as follows:

Let $I$ be an arbitrary satellite image of size $W \times H \times D$, where $W$ and $H$ are the width and height of the images in pixels, respectively, while $D$ is the depth of the images (i.e., the number of features per pixel). The objective is to automatically create a binary mask $M$ represented by a matrix of size $W \times H$ associated with $I$, where the value 1 in a cell indicates the associated pixel contains cultivated area. In contrast, 0 is related to any non-cultivated area.

\subsubsection{Explainability Statement}
We aim to help users fully understand how the model achieves effective segmentation by providing them with visual explanations of model predictions. 
From an XAI perspective, the problem can be formulated as follows:

Given an image $I$ and its binary mask $M$, we want to produce a saliency map (or heat map) $S$ of size $W \times H$ to highlight the regions of $I$ that are important for the model's prediction. 
Each element $s_i$ of $S$ is the importance score associated with pixel $p_i$ of image $I$. 
Each value $s_i$ represents the influence $p_i$ on the prediction of cultivated area. Visualizing the saliency map eases the interpretation of how the model arrives at its decisions. 

\subsection{Models}
In this study, we compare the well-known U-Net~\cite{unet} with a modified version~\cite{ukan} which integrates KAN~\cite{kan} layers into the architecture. 
In the following, we first outline the U-Net architecture. 
We then outline the KAN neural networks and, finally, its integration into the U-KAN architecture.

\subsubsection{U-Net}
\begin{figure}[htb]
    \centering
    \includegraphics[width=\linewidth]{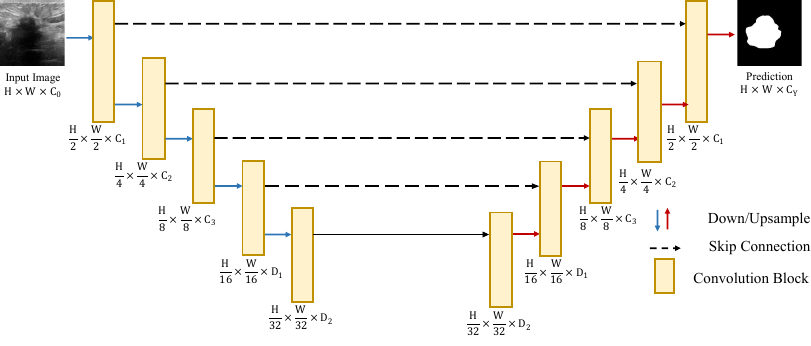}
    \caption{U-Net architecture~\cite{unet}. It is characterized by a U-shape structure with a contracting path (left side) to capture context and a symmetric expanding path (right side) to enable precise localization.
    } 
    \label{fig:unet}
\end{figure}

U-Net is a convolutional neural network architecture designed primarily for biomedical image segmentation~\cite{unet}. Its architecture is characterized by a U-shaped structure (as seen in \Cref{fig:unet}), with a contracting path to capture context and a symmetric expanding path to enable precise localization. The contracting path consists of repeated application of convolutions followed by max-pooling operations, while the expanding path involves upsampling and convolutional layers to reconstruct the image resolution. This design allows U-Net to effectively learn from relatively few training images and produce high-quality segmentations, making it a popular choice for segmentation beyond medical imaging.

\subsubsection{KAN}
Kolmogorov–Arnold Networks~\cite{kan} (KANs) are a novel type of neural network inspired by the Kolmogorov-Arnold representation theorem~\cite{kolmogorov1956representation,arnold1957functions}, which states that every multivariate continuous function $f: [0,1]^{n}\to \mathbb {R}$ can be represented as a superposition of the two-argument addition of continuous functions of one variable:
\begin{equation}
    f(x) = f(x_1, x_2, ..., x_n) = \sum_{q=0}^{2n} \Phi_q \Big(\sum_{p = 1}^n \phi_{q,p} (x_p)\Big)
\end{equation}
where $\phi _{q,p}: [0,1]\to \mathbb {R}$ and $\Phi_{q}: \mathbb {R} \to \mathbb {R}$.
Unlike traditional Multi-Layer Perceptrons (MLPs) that have fixed activation functions on nodes, KANs employ learnable activation functions on edges. This is achieved by replacing every linear weight parameter with a univariate function parameterized as a spline. The activations change step-by-step to better approximate the desired target during the training, and KANs offer the possibility of visualizing the learnable activation functions.
In this way, KANs can be more transparent and efficient in learning more complex relations than MLPs, offering promising alternatives to traditional deep learning models. The learned activations can be low-cost functions (e.g., constant or linear) when there is no need for complex non-linearities. This also grants the possibility of understanding the salient parts of the input.

\subsubsection{U-KAN}
\begin{figure}[htb]
    \centering
    \includegraphics[width=\linewidth]{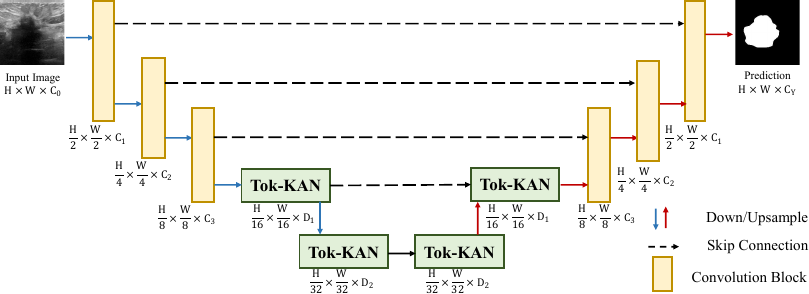}
    \caption{U-KAN~\cite{ukan}. It has the same U-shape as U-Net, but the deepest layers are implemented as Tok-KAN blocks. Tok-KAN blocks are composed of a Tokenization layer, a KAN layer, a Downsampling layer, and 
 a Normalization layer.}
    \label{fig:ukan}
\end{figure}

U-KAN~\cite{ukan} proposes to implement the deepest layer of the U-Net using KANs. These layers are composed of a tokenization layer, a KAN layer, a downsampling layer, and a final normalization layer. In \Cref{fig:ukan}, it is possible to see the key difference stands only in how the deepest representations are processed by the network. The main features of the U-Net, such as downsampling and skip-connection, remain invariant, sharing the same benefits. The modification in the encoder's last layers and the decoder's firsts allows the network to learn custom activation functions instead of fixed ones, potentially improving the representativity of the embeddings and reducing the required computational resources by learning simple activations when needed. %

\section{Experimental Setup}
\label{sec:setup}
This section describes the adopted dataset, the experimental setting, and the adopted evaluation metrics for the crop segmentation performance and quality of the derived explanations.

\subsection{Dataset}
We employed the South Africa Crop Type dataset~\cite{sa_crop_type_dataset}, which contains images taken from both Sentinel-2 and Sentinel-1 covering a wide region of South Africa. 
The dataset includes small and irregular-shaped~\cite{ghana_sudan_crops} crop fields, making it more challenging, 
and provides higher resolution imagery (of size $256 \times 256$) than other datasets covering the area.
The annotations contain the mask of the areas covered by a certain crop. For our analysis, we limit the scope to distinguish between cultivated and non-cultivated areas. 
We analyzed the results using both types of imagery from Sentinel-2 and Sentinel-1.

Sentinel-1~\cite{sentinel1} is a satellite mission under the Copernicus program, comprising two identical satellites equipped with C-band Synthetic Aperture Radar (SAR). It provides all-weather, day-and-night radar imaging. %
The satellite can work in both single-polarization and double-polarization modes. On land, it mainly works collecting VV and VH polarizations%

Sentinel-2~\cite{sentinel2} is part of the Copernicus program and consists of two satellites. These satellites carry high-resolution multispectral imaging instruments with 13 spectral bands ranging from Ultra-Blue, Visible, Near Infrared (NIR), and Short Wave infrared (SWIR). %
It is particularly sensitive to vegetation due to the presence of instruments that work in the infrared spectrum. %

While Sentinel-1 images can cover different atmospheric situations due to the radiometric nature of the imagery, Sentinel-2 is affected by clouds and similar atmospheric disturbances. Since the provided cloud masks are often not accurate, we computed the masks with the s2cloudless algorithm~\cite{s2cloudless}. We exclude low-quality Sentinel-2 images with a strong overlap between the cloud mask and the areas containing crops (intersection over 0.7).

\begin{figure}[htb]
    \centering
    \includegraphics[width=0.9\linewidth]{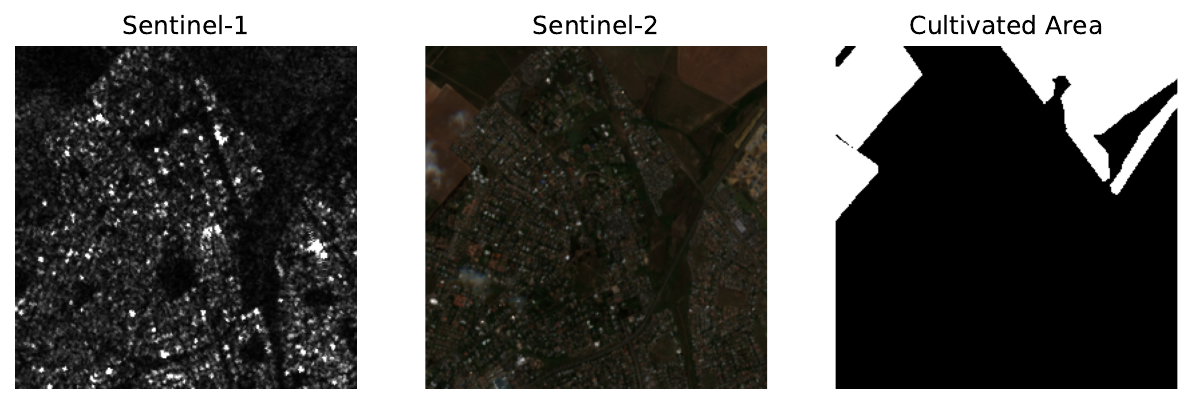}
    \caption{South Africa Crop Type dataset example. Sentinel-1 shows the grayscale of the VV polarization, while Sentinel-2 shows the RGB of the same area. The mask shows the cultivated area in white.}
    \label{fig:sa_dataset}
\end{figure}

Since no splits were provided, we randomly divided the dataset into a training set containing 2019 training, 267 validation, and 364 test samples. The three splits are similar ($p > 0.9$) according to the chi-square test when measuring the class frequencies. In this way, we created two datasets with three splits each due to the fact the dates of Sentinel-1 and Sentinel-2 do not exactly match because of the different revisit times. \Cref{fig:sa_dataset}, show a sample from the test set in both Sentinel-1 VV and Sentinel-2 RGB.

\subsection{Experimental Setting}

\subsubsection{Crop Field Segmentation}
The images have size $256 \times 256 \times 2$ for Sentinel-1 data and shape $256 \times 256 \times 12$ for Sentinel-2.
We train all networks with an AdamW optimizer and a learning rate scheduler with a reduction on plateau of factor 0.2 and patience of 5. 
The initial learning rate was set to 1e-4, and the batch size to 16. 
We trained the models for 60 epochs. 
We apply random horizontal and vertical flipping as augmentations. 
The loss function is generalized dice loss~\cite{gdice}, which takes into account the class imbalance in the images. 
We compared the two networks with the same encoder (and so decoder) embedding sizes to better understand how they exploit the same representation space. 

We evaluated the networks under Intersection-Over-Union (IoU), F1-Score (F1), Precision (Prec), and Recall (Rec) for the positive class. 
We also evaluated GFLOPs to measure the efficiency of the network.

\subsubsection{Explainability}
We use Grad-CAM~\cite{selvaraju2016grad} as a visual post-hoc explanation method for this task because of its proven effectiveness in previous XAI studies in remote sensing~\cite{kakogeorgiou2021evaluating}. Grad-CAM is particularly valuable because it helps answer the critical question, \textit{``Where does the model focus when segmenting crops?''}.

For each image, we generate a single saliency map to quantify the influence of each pixel on the prediction of the positive class (i.e., cultivated area)  made by a model (i.e., U-NET or U-KAN).
In GradCAM, the generation process first involves computing the gradients of the positive class score with respect to feature maps of a selected convolutional layer. These gradients are then averaged globally to obtain the importance weights for each feature map. Next, a weighted sum of these feature maps is performed using the calculated weights. This yields a coarse location map that highlights the regions of the input image that are most influential in model segmentation decisions. Finally, we apply a ReLU activation to this weighted sum to ensure that only positive influences are considered, producing the final Grad-CAM heat map. 
In our experiments, we use Sentinel-2 data, which provides multispectral images over 12 channels, and generate the explanations for the test set images.

We assessed the \textit{Plausibility}, \textit{Sufficiency}, and \textit{Per-channel Relevance} of the generated Grad-CAM heatmaps. 
Below, we provide a detailed description of each metric.

\paragraph{Plausibility}\label{par:plausibility}
Plausibility refers to the degree to which the explanations align with human understanding and domain-specific knowledge~\cite{doshi2017towards, samek2017explainable, kakogeorgiou2021evaluating}. This is crucial in ensuring that the models not only perform well but are also aligned with human expectations and knowledge. In this study, we want to assess to which extent each obtained heat map is aligned with the relative ground truth. 
We evaluate the plausibility of saliency maps by calculating each metric (IoU, F1, Prec, Rec) between the generated saliency map and the corresponding ground truth mask. 

Since our saliency maps provide continuous explanations where each pixel has a value of importance, we established a threshold of importance to define which pixels are considered important for the segmentation with Otsu method~\cite{otsu_threshold}. %
This method segments the saliency map into distinct regions, creating a binary mask directly comparable to the binary truth mask. 

\paragraph{Sufficiency}
Sufficiency is an aspect of faithfulness, evaluating whether an explanation indeed captures the important factors contributing to the segmentation and, therefore, is sufficient. \cite{ribeiro2016should, lundberg2017unified}. 

To assess the sufficiency of an explanation, we preserve only the important pixels identified by the explanation and mask the others. We then evaluate the performance metrics (IoU, F1-score, Precision, and Recall) for the positive class on this altered image. Sufficiency is computed as the change in metrics between the original and altered images. A smaller drop in performance indicates a more sufficient explanation. In this case, we also use the Otsu method to threshold the binary saliency maps.

\paragraph{Per-channel Relevance}
Another important aspect of standard XAI evaluation is the variation in performance metrics when the input image is perturbed. Occlusion sensitivity \cite{borys2023explainable} is a method that involves systematically masking parts of the input image using a sliding window and measuring the change in the model's output. This technique identifies critical image regions for the model's predictions, offering insights into its reasoning and the faithfulness of its explanations.

In our specific situation, we apply the idea of occlusion not to parts of the image but to entire channels. This approach aligns better with the nature of our data, where each pixel carries its own importance and classification. 
By occluding an entire channel, we can systematically evaluate how the absence of specific channels affects the model's explanation, providing clearer insights into the role each channel plays in the classification process.

In our tests, we occlude one channel at a time and calculate the saliency map. 
Then, we measure the IoU between the saliency map obtained by occluding one channel and the saliency map obtained using all channels. 

\section{Experimental Results}
\label{sec:results}
In this section, we present the results obtained for the analyzed dataset. 
First, we outline crop segmentation performance, addressing our research question RQ1. 
We then analyze the explanations of U-Net and U-KAN predictions from the qualitative and quantitative perspectives, addressing RQ2.

\subsection{Task Performances}
In \Cref{tab:results}, we report the results obtained when employing U-Net and U-KAN on Sentinel-1 and Sentinel-2 data.
U-KAN provides the best performance in terms of IoU on Sentinel-2, proving its adaptability in dealing with complex relations. On Sentinel-1, U-KAN gets comparable performance in terms of IoU to U-Net.
In terms of precision, the KAN variant is more performant, scoring $\approx +3\%$ than U-Net. 
Although Sentinel-1 imagery is less affected by atmospheric events, Sentinel-2 bands provide a better understanding of the area with both networks.
U-KAN proves to be more computationally efficient than a standard U-Net when looking at GFLOPS: it consumes half the one of U-Net. 

The KAN variant proves to be a preferable solution in both cases, providing better or comparable performance in fewer GFLOPs. Moreover, it proves to be more precise in any case.

\input{Tables/table_task_performances}
\subsection{Analysis of explanations}
We analyze the explainability results of U-KAN and U-Net networks on the Sentinel-2 dataset through qualitative and quantitative assessments.

\subsubsection{Qualitative Evaluation}
We examine the saliency maps obtained from both networks. This analysis provides insight into the areas where each model focuses its attention, offering a deeper understanding of their segmentation behaviors.

Figure~\ref{fig:saliency_comp} shows examples of saliency maps generated by the U-Net and U-KAN models. 
The red pixels indicate the points where the network's attention is most focused, highlighting the differences in the behavior of the two models. 
Figure~\ref{fig:saliency_comp} (c) reveals that the U-Net model focuses on a significantly larger area than the U-KAN model.
This observation suggests that, regardless of the effectiveness of the segmentation task, the U-Net model tends to distribute its focus over larger regions.
In contrast, the U-KAN model has an interesting feature in its approach to the segmentation task. The network focuses predominantly on the boundaries of cultivated areas rather than within the areas themselves. This focus on boundaries suggests that the U-KAN model prioritizes delineating the edges of areas of interest. This last feature opens up the potential use of U-KAN networks in boundary delimitation and mapping tasks, where precise edge detection and delineation are crucial\cite{Xie_2015_ICCV, mahdianpari2018very}.
These considerations generally apply to the images in the dataset. We included multiple examples of saliency maps in our repository.

\begin{figure}[htb]
    \centering
    \subfloat[RGB and GT]{\includegraphics[width=0.18\linewidth]{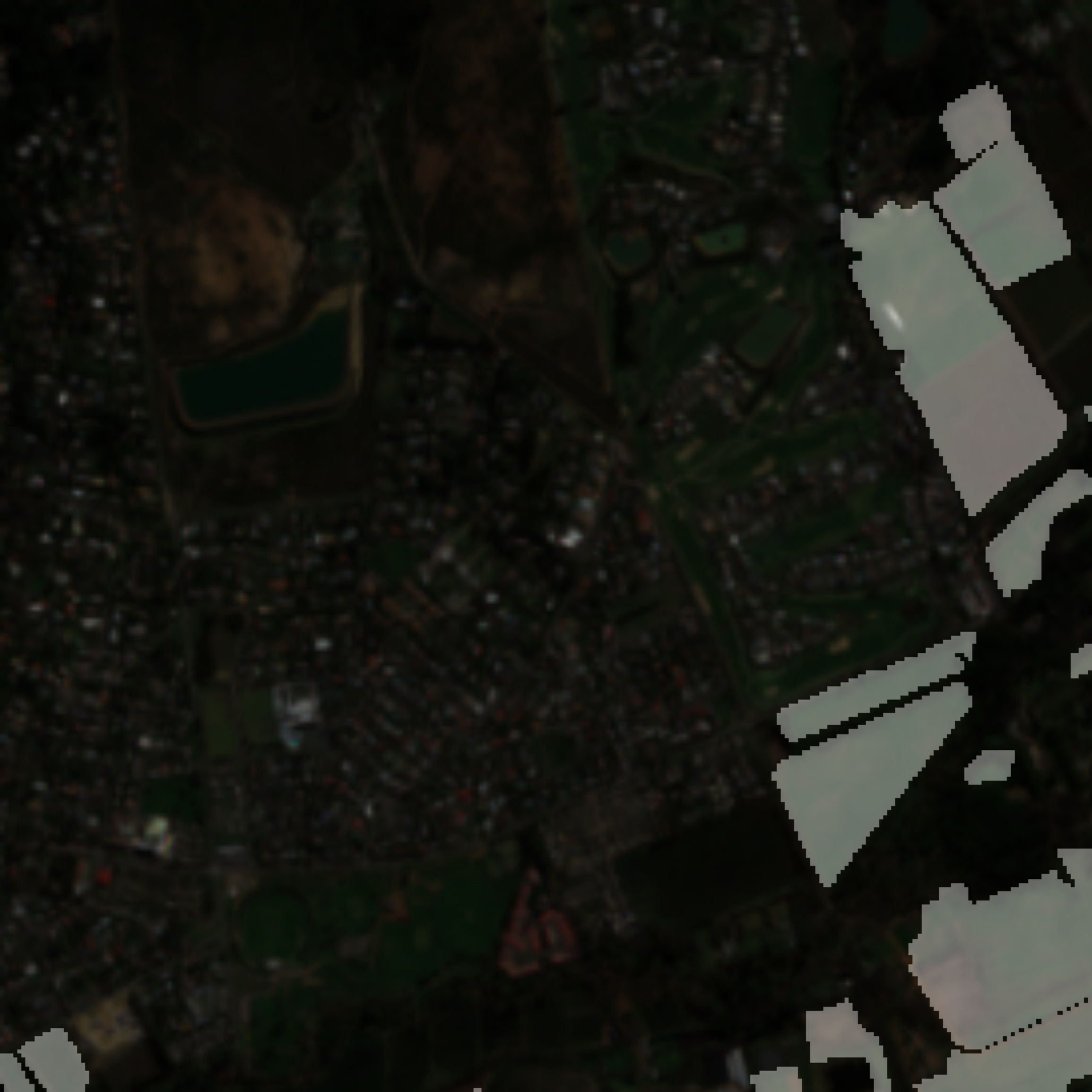}} \hfill
    \subfloat[All Channels]{\includegraphics[width=0.18\linewidth]{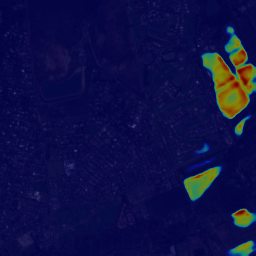}} \hfill
    \subfloat[No B01]{\includegraphics[width=0.18\linewidth]{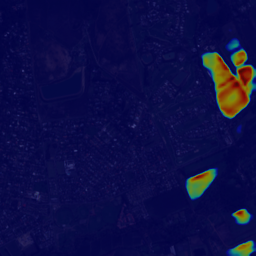}} \hfill
    \subfloat[No B06]{\includegraphics[width=0.18\linewidth]{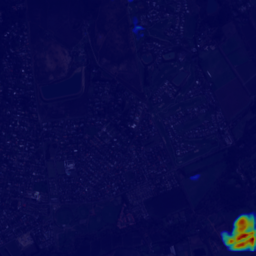}} \hfill
    \subfloat[No B11]{\includegraphics[width=0.18\linewidth]{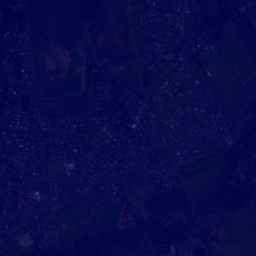}}
    \caption{Per-channel relevance examples of U-KAN. The figure shows the ground truth over the original RGB image (a) and the saliency map of all 12 channels (b). Images (c), (d), and (e) display saliency maps generated by obscuring channels corresponding to B01, B06, and B11, respectively.}
    \label{fig:relevance channels}
\end{figure}

Figure~\ref{fig:relevance channels} illustrates the change in saliency maps when specific channels related to bands B01, B06, and B11 are obscured for the U-KAN. 
From Figure~\ref{fig:relevance channels}(c), we can infer that by obscuring the channel relative to band B01, the resulting saliency map is very similar to the original one (Figure~\ref{fig:relevance channels}(b)), indicating that this channel has little impact on model focus. 
When the channel corresponding to band B06 is obscured, we observe a change in the saliency map, suggesting an influence of this channel on network focus. 
The most significant change occurs when we obscure the channel relative to B11 (Figure~\ref{fig:relevance channels}(e)). 
The resulting saliency map shows that the model does not focus on any specific area, implying that the network cannot detect segmentation traits without this channel. 
This result highlights the critical importance of band B11 in the model's ability to perform the segmentation task.
Similar observations apply to the U-Net. 
We further investigate the impact of channels on saliency maps in the following quantitive analysis.

\input{Tables/table_plausibility}

\subsubsection{Quantitative Evaluation}
Table \ref{tab:plausibility} presents the evaluation results of \textit{Plausibility}. 
Regarding the plausibility of explanations, U-KAN demonstrates higher IoU and Precision than its competitor, U-Net. 
This indicates that U-KAN provides more accurate and reliable explanations, aligning more closely with human understanding. %
On the other hand, U-Net has a higher Recall and F1 score of plausibility. This could be because U-Net is more sensitive and captures more features in saliency maps, although it includes more false positives. 

\input{Tables/table_sufficicency}

In Table \ref{tab:sufficiency}, we present the results assessing the \textit{Sufficiency} of explanations. 
Sufficiency is quantified as the difference in metrics between masked images and their original counterparts. 
An interesting observation is the variation in the Precision metric. While there is a decrease in other metrics, which aligns with removing less critical pixels, there is an improvement in Precision for both U-KAN and U-Net.
This increase in precision is noteworthy because by excluding less important pixels, both networks demonstrate an enhanced ability to delineate pixels belonging to crop class.

\input{Tables/table_relevance}

Table \ref{tab:relevance} reports the \textit{Per-channel relevance} results. For each model, we report the IoU score between the saliency map of all channels and the one obtained by obscuring the specific channel related to the band. 
In this context, a lower IoU signifies the higher importance of a channel. 
Specifically, if removing a channel yields a lower or zero IoU, it indicates that the channel plays an important role in the final segmentation task. 
For both U-KAN and U-Net models, we obtain that the channels corresponding to bands B05 (705 nm - Red Edge), B8A (865 nm - Narrow Near-Infrared), and B11 (1610 nm - Shortwave Infrared 1) are identified as the most significant for crop segmentation task due to their specific sensitivities. 
Specifically, B05 and B8A are sensitive to chlorophyll content and vegetation biomass and B11 to moisture content in soil and vegetation \cite{broge2001comparing, van2014potential, castaldi2019evaluating}.
This quantitative evaluation across all test samples aligns with the qualitative results displayed in Figure~\ref{fig:relevance channels}, confirming the discussed insights. 
Analyzing the relevance of each channel in Sentinel-2 images opens the possibility of reducing the number of channels used by focusing on only the most important ones. 
This optimization can enhance efficiency and reduce computational costs while maintaining the quality of the analysis.

\subsection{Analysis of the trained models}
As previously mentioned, KANs are designed with a level of interpretability through the possibility of visualizing learnable activation functions. Although our main goal is to explore the post-hoc explainability of the U-KAN network with respect to the U-Net, we also sought to analyze the behavior of learnable activation functions in a decoding layer. This is also particularly relevant when analyzing the resource consumption of the network since the learned function can be simple to compute (e.g., linear) or complex when necessary, potentially helping both GFLOPs and performance, as shown before. 

\begin{figure}[htb]
    \centering
    \includegraphics[width=0.8\linewidth]{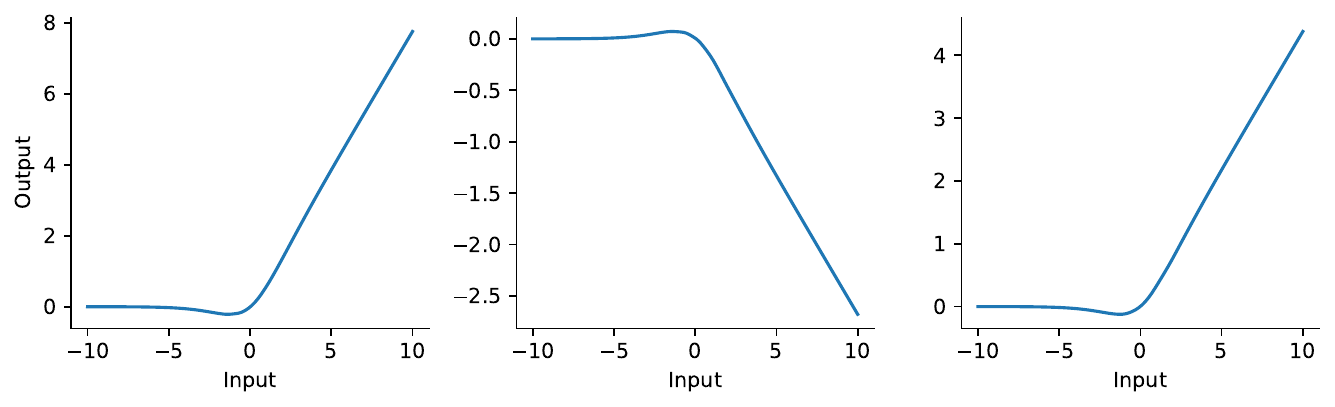}
    \caption{Activation functions for the first element of the embeddings of the KAN decoder layer.}
    \label{fig:activations_0}
\end{figure}

In \Cref{fig:activations_0}, we report the learned activations for an element of the embeddings in a decoder layer. 
We can see substantial differences between the base function (SiLU) and ReLU, commonly employed by U-Net. U-KAN effectively represented more complex relations in the deep embeddings. The second activation is reversed along the y-axis compared to SiLU. The first and the third are similar but have a different slope (the first function is steeper).

Moreover, each element of the embedding learns different activations with different complexity.
Some of them could be constant activations for certain elements in the embeddings. %
The learned functions with a variance < 1 are $\approx 26\%$ the total, while the ones with a variance < 0.1 are $\approx 8\%$. This indicates the irrelevant parts of the embedding because every input is mapped to the same value every time.

\section{Conclusions}
\label{sec:conclusions}
In this work, we have shown how the new KANs can improve well-known architecture applied to the agricultural field, particularly in terms of efficiency, using only half the resources of full CNN architecture. 
Our study indicates that U-KAN offers superior performance by achieving higher precision and IoU scores than U-Net. %
The explainability analysis also reveals two significant insights. First, the U-KAN network's emphasis on boundary details makes it particularly effective for tasks such as boundary delimitation and mapping.
Second, not all the channels are useful for the segmentation task. So, users can decide to rely only on the most important reducing computational costs of the models.
In future work, we intend to implement the insights from our network explainability analysis to enhance performance and reduce computational costs. %

\bibliographystyle{splncs04}
\bibliography{bibliography}
\end{document}

%% file: Tables/table_task_performances.tex
\begin{table}[htb]    
    \caption{Results of U-Net and U-KAN using Sentinel-1 (S1) and Sentinel-2 (S2) data.}
    \label{tab:results}
    \centering
    \begin{tabular}{l|lc|cccc} \toprule
                         & Model & GFLOPs $\downarrow$ & IoU $\uparrow$ & F1 $\uparrow$ & Prec $\uparrow$ & Rec $\uparrow$ \\ \midrule
    \multirow{2}{*}{S1}  &  U-Net & 79.89  & \textbf{65.59} & \textbf{79.21} & 74.56 & \textbf{85.56} \\
                         &  U-KAN & \textbf{44.89}  & 65.36 & 79.03 & \textbf{77.40} & 77.50 \\ \midrule
    \multirow{2}{*}{S2}  &  U-Net & 80.65  & 72.95 & 84.35 & 72.57 & \textbf{94.33} \\
                         &  U-KAN & \textbf{45.65}  & \textbf{74.82} & \textbf{85.59} & \textbf{75.24} & 93.31 \\
    \bottomrule
    \end{tabular}
\end{table}

%% file: Tables/table_plausibility.tex
\begin{table}[htb]
    \caption{Plausibility of Explanations for Sentinel-2.}
    \centering
    \begin{tabular}{l|cccc}
        \toprule
         Model & IoU & F1 & Prec & Rec \\
         \midrule

         U-KAN & \textbf{73.52} & 80.77 &  \textbf{84.49 }& 77.74\\
         U-Net& 68.19 & \textbf{84.50} & 82.23& \textbf{87.13}\\
        \bottomrule
    \end{tabular}

    \label{tab:plausibility}
\end{table}

%% file: Tables/table_sufficicency.tex
\begin{table}[ht]
    \caption{Sufficiency of Explanations for Sentinel-2.}
    \centering
    \begin{tabular}{l | c | c | c | c}
        \toprule
        Model &  IoU & F1 & Prec & Rec \\
        \midrule
         U-KAN %
                                & 67.94 (-6.88) & 80.59 (-5.0) & 84.46 (+9.22) & 77.45 (-15.86) \\
         U-Net %
                                & 68.85 (-4.1) & 81.26 (-3.09) & 84.10 (+11.53) & 78.95 (-15.38) \\
        \bottomrule
    \end{tabular}
    \label{tab:sufficiency}
\end{table}

%% file: Tables/table_relevance.tex
\begin{table}[ht]
    \caption{Per-channel relevance (lower is better) based on IoU for Sentinel-2.}
        \label{tab:relevance}

    \centering
    \begin{tabular}{@{}c|cccccccccccc@{}}
    \toprule
                   & B01   & B02   & B03   & B04   & B05           & B06   & B07   & B08   & B8A           & B09   & B11           & B12   \\ \midrule
    U-KAN & 72.92 & 74.62 & 46.90 & 45.90 & \textbf{0.11} & 47.49 & 38.51 & 15.61 & \textbf{0.00} & 42.48 & \textbf{0.19} & 47.03 \\
    U-Net & 68.08 & 68.90 & 46.24 & 46.26 & \textbf{0.13} & 47.19 & 36.91 & 15.54 & \textbf{0.00} & 42.55 & \textbf{0.18} & 48.30 \\ \bottomrule
    \end{tabular}

\end{table}

%% file: main.bbl
\begin{thebibliography}{10}
\providecommand{\url}[1]{\texttt{#1}}
\providecommand{\urlprefix}{URL }
\providecommand{\doi}[1]{https://doi.org/#1}

\bibitem{adadi2018peeking}
Adadi, A., Berrada, M.: Peeking inside the black-box: a survey on explainable artificial intelligence (xai). IEEE access  \textbf{6},  52138--52160 (2018)

\bibitem{arnold1957functions}
Arnold, V.: On functions of three variables. In: Proceedings of the USSR Academy of Sciences. pp. 678--681 (1957)

\bibitem{arrieta2020explainable}
Arrieta, A.B., D{\'\i}az-Rodr{\'\i}guez, N., Del~Ser, J., Bennetot, A., Tabik, S., Barbado, A., Garc{\'\i}a, S., Gil-L{\'o}pez, S., Molina, D., Benjamins, R., et~al.: Explainable artificial intelligence (xai): Concepts, taxonomies, opportunities and challenges toward responsible ai. Information fusion  \textbf{58},  82--115 (2020)

\bibitem{unet_crop_segmentation}
Ayushi, Buttar, P.K.: Satellite imagery analysis for crop type segmentation using u-net architecture. Procedia Computer Science  \textbf{235},  3418–3427 (2024). \doi{10.1016/j.procs.2024.04.322}, \url{http://dx.doi.org/10.1016/j.procs.2024.04.322}

\bibitem{spectral_inputs_crop}
Bauer, M.: Spectral inputs to crop identification and condition assessment. Proceedings of the IEEE  \textbf{73}(6),  1071--1085 (1985). \doi{10.1109/PROC.1985.13238}

\bibitem{biotic_stress_review}
Behmann, J., Mahlein, A.K., Rumpf, T., Römer, C., Plümer, L.: A review of advanced machine learning methods for the detection of biotic stress in precision crop protection. Precision Agriculture  \textbf{16}(3),  239–260 (Aug 2014). \doi{10.1007/s11119-014-9372-7}, \url{http://dx.doi.org/10.1007/s11119-014-9372-7}

\bibitem{precision_farming_impact}
Belal, A.A., EL-Ramady, H., Jalhoum, M., Gad, A., Mohamed, E.S.: Precision Farming Technologies to Increase Soil and Crop Productivity, p. 117–154. Springer International Publishing (2021). \doi{10.1007/978-3-030-78574-1_6}, \url{http://dx.doi.org/10.1007/978-3-030-78574-1_6}

\bibitem{borys2023explainable}
Borys, K., Schmitt, Y.A., Nauta, M., Seifert, C., Kr{\"a}mer, N., Friedrich, C.M., Nensa, F.: Explainable ai in medical imaging: An overview for clinical practitioners--saliency-based xai approaches. European journal of radiology  \textbf{162},  110787 (2023)

\bibitem{broge2001comparing}
Broge, N.H., Leblanc, E.: Comparing prediction power and stability of broadband and hyperspectral vegetation indices for estimation of green leaf area index and canopy chlorophyll density. Remote sensing of environment  \textbf{76}(2),  156--172 (2001)

\bibitem{crop_type_segmentation}
Buttar, P.K., Sachan, M.K.: Semantic segmentation of satellite images for crop type identification in smallholder farms. The Journal of Supercomputing  \textbf{80}(2),  1367–1395 (Jul 2023). \doi{10.1007/s11227-023-05533-4}, \url{http://dx.doi.org/10.1007/s11227-023-05533-4}

\bibitem{castaldi2019evaluating}
Castaldi, F., Hueni, A., Chabrillat, S., Ward, K., Buttafuoco, G., Bomans, B., Vreys, K., Brell, M., van Wesemael, B.: Evaluating the capability of the sentinel 2 data for soil organic carbon prediction in croplands. ISPRS Journal of Photogrammetry and Remote Sensing  \textbf{147},  267--282 (2019)

\bibitem{change_detection_reality}
Corley, I., Robinson, C., Ortiz, A.: A change detection reality check. arXiv preprint arXiv:2402.06994  (2024)

\bibitem{doshi2017towards}
Doshi-Velez, F., Kim, B.: Towards a rigorous science of interpretable machine learning. arXiv preprint arXiv:1702.08608  (2017)

\bibitem{sentinel2}
Drusch, M., Del~Bello, U., Carlier, S., Colin, O., Fernandez, V., Gascon, F., Hoersch, B., Isola, C., Laberinti, P., Martimort, P., et~al.: Sentinel-2: Esa's optical high-resolution mission for gmes operational services. Remote sensing of Environment  \textbf{120},  25--36 (2012)

\bibitem{gevaert2022explainable}
Gevaert, C.M.: Explainable ai for earth observation: A review including societal and regulatory perspectives. International Journal of Applied Earth Observation and Geoinformation  \textbf{112},  102869 (2022)

\bibitem{huang2021better}
Huang, X., Sun, Y., Feng, S., Ye, Y., Li, X.: Better visual interpretation for remote sensing scene classification. IEEE Geoscience and Remote Sensing Letters  \textbf{19}, ~1--5 (2021)

\bibitem{hung2021integrating}
Hung, S.C., Wu, H.C., Tseng, M.H.: Integrating image quality enhancement methods and deep learning techniques for remote sensing scene classification. Applied Sciences  \textbf{11}(24),  11659 (2021)

\bibitem{kakogeorgiou2021evaluating}
Kakogeorgiou, I., Karantzalos, K.: Evaluating explainable artificial intelligence methods for multi-label deep learning classification tasks in remote sensing. International Journal of Applied Earth Observation and Geoinformation  \textbf{103},  102520 (2021)

\bibitem{disaster_management_review}
Khan, S.M., Shafi, I., Butt, W.H., Diez, I.d.l.T., Flores, M.A.L., Gal{\'a}n, J.C., Ashraf, I.: A systematic review of disaster management systems: approaches, challenges, and future directions. Land  \textbf{12}(8), ~1514 (2023)

\bibitem{kolmogorov1956representation}
Kolmogorov, A.N.: On the representation of continuous functions of several variables by superpositions of continuous functions of a smaller number of variables. In: Proceedings of the USSR Academy of Sciences (1956)

\bibitem{levering2020interpretable}
Levering, A., Marcos, D., Lobry, S., Tuia, D.: Interpretable scenicness from sentinel-2 imagery. In: IGARSS 2020-2020 IEEE International Geoscience and Remote Sensing Symposium. pp. 3983--3986. IEEE (2020)

\bibitem{ukan}
Li, C., Liu, X., Li, W., Wang, C., Liu, H., Yuan, Y.: U-kan makes strong backbone for medical image segmentation and generation. arXiv preprint arXiv:2406.02918  (2024)

\bibitem{kan}
Liu, Z., Wang, Y., Vaidya, S., Ruehle, F., Halverson, J., Soljačić, M., Hou, T.Y., Tegmark, M.: Kan: Kolmogorov-arnold networks (2024), \url{https://arxiv.org/abs/2404.19756}

\bibitem{lundberg2017unified}
Lundberg, S.M., Lee, S.I.: A unified approach to interpreting model predictions. Advances in neural information processing systems  \textbf{30} (2017)

\bibitem{ghana_sudan_crops}
M~Rustowicz, R., Cheong, R., Wang, L., Ermon, S., Burke, M., Lobell, D.: Semantic segmentation of crop type in africa: A novel dataset and analysis of deep learning methods. In: Proceedings of the IEEE/CVF Conference on Computer Vision and Pattern Recognition (CVPR) Workshops (June 2019)

\bibitem{maddy2021miidaps}
Maddy, E.S., Boukabara, S.A.: Miidaps-ai: An explainable machine-learning algorithm for infrared and microwave remote sensing and data assimilation preprocessing-application to leo and geo sensors. IEEE Journal of Selected Topics in Applied Earth Observations and Remote Sensing  \textbf{14},  8566--8576 (2021)

\bibitem{mahdianpari2018very}
Mahdianpari, M., Salehi, B., Rezaee, M., Mohammadimanesh, F., Zhang, Y.: Very deep convolutional neural networks for complex land cover mapping using multispectral remote sensing imagery. Remote Sensing  \textbf{10}(7), ~1119 (2018)

\bibitem{martinez2020crop}
Mart{\'\i}nez-Ferrer, L., Piles, M., Camps-Valls, G.: Crop yield estimation and interpretability with gaussian processes. IEEE Geoscience and Remote Sensing Letters  \textbf{18}(12),  2043--2047 (2020)

\bibitem{mateo2021learning}
Mateo-Sanchis, A., Piles, M., Amor{\'o}s-L{\'o}pez, J., Mu{\~n}oz-Mar{\'\i}, J., Adsuara, J.E., Moreno-Mart{\'\i}nez, {\'A}., Camps-Valls, G.: Learning main drivers of crop progress and failure in europe with interpretable machine learning. International Journal of Applied Earth Observation and Geoinformation  \textbf{104},  102574 (2021)

\bibitem{van2014potential}
Van~der Meer, F., Van~der Werff, H., Van~Ruitenbeek, F.: Potential of esa's sentinel-2 for geological applications. Remote sensing of environment  \textbf{148},  124--133 (2014)

\bibitem{land_cover_review}
Moharram, M.A., Sundaram, D.M.: Land use and land cover classification with hyperspectral data: A comprehensive review of methods, challenges and future directions. Neurocomputing  \textbf{536},  90–113 (Jun 2023). \doi{10.1016/j.neucom.2023.03.025}, \url{http://dx.doi.org/10.1016/j.neucom.2023.03.025}

\bibitem{molnar2020interpretable}
Molnar, C.: Interpretable machine learning. Lulu. com (2020)

\bibitem{remote_sensing_agriculture_review}
Mulla, D.J.: Twenty five years of remote sensing in precision agriculture: Key advances and remaining knowledge gaps. Biosystems Engineering  \textbf{114}(4),  358–371 (Apr 2013). \doi{10.1016/j.biosystemseng.2012.08.009}, \url{http://dx.doi.org/10.1016/j.biosystemseng.2012.08.009}

\bibitem{otsu_threshold}
Otsu, N.: A threshold selection method from gray-level histograms. IEEE Transactions on Systems, Man, and Cybernetics  \textbf{9}(1),  62--66 (1979). \doi{10.1109/TSMC.1979.4310076}

\bibitem{ribeiro2016should}
Ribeiro, M.T., Singh, S., Guestrin, C.: " why should i trust you?" explaining the predictions of any classifier. In: Proceedings of the 22nd ACM SIGKDD international conference on knowledge discovery and data mining. pp. 1135--1144 (2016)

\bibitem{unet}
Ronneberger, O., Fischer, P., Brox, T.: U-net: Convolutional networks for biomedical image segmentation. In: Medical image computing and computer-assisted intervention--MICCAI 2015: 18th international conference, Munich, Germany, October 5-9, 2015, proceedings, part III 18. pp. 234--241. Springer (2015)

\bibitem{samek2017explainable}
Samek, W., Wiegand, T., M{\"u}ller, K.R.: Explainable artificial intelligence: Understanding, visualizing and interpreting deep learning models. arXiv preprint arXiv:1708.08296  (2017)

\bibitem{selvaraju2016grad}
Selvaraju, R.R., Das, A., Vedantam, R., Cogswell, M., Parikh, D., Batra, D.: Grad-cam: Why did you say that? arXiv preprint arXiv:1611.07450  (2016)

\bibitem{s2cloudless}
Skakun, S., Wevers, J., Brockmann, C., Doxani, G., Aleksandrov, M., Batič, M., Frantz, D., Gascon, F., Gómez-Chova, L., Hagolle, O., López-Puigdollers, D., Louis, J., Lubej, M., Mateo-García, G., Osman, J., Peressutti, D., Pflug, B., Puc, J., Richter, R., Roger, J.C., Scaramuzza, P., Vermote, E., Vesel, N., Zupanc, A., Žust, L.: Cloud mask intercomparison exercise (cmix): An evaluation of cloud masking algorithms for landsat 8 and sentinel-2. Remote Sensing of Environment  \textbf{274},  112990 (Jun 2022). \doi{10.1016/j.rse.2022.112990}, \url{http://dx.doi.org/10.1016/j.rse.2022.112990}

\bibitem{stomberg2021jungle}
Stomberg, T., Weber, I., Schmitt, M., Roscher, R.: Jungle-net: Using explainable machine learning to gain new insights into the appearance of wilderness in satellite imagery. ISPRS Annals of the Photogrammetry, Remote Sensing and Spatial Information Sciences  \textbf{3},  317--324 (2021)

\bibitem{gdice}
Sudre, C.H., Li, W., Vercauteren, T., Ourselin, S., Jorge~Cardoso, M.: Generalised Dice Overlap as a Deep Learning Loss Function for Highly Unbalanced Segmentations, p. 240–248. Springer International Publishing (2017). \doi{10.1007/978-3-319-67558-9_28}, \url{http://dx.doi.org/10.1007/978-3-319-67558-9_28}

\bibitem{temenos2023interpretable}
Temenos, A., Temenos, N., Kaselimi, M., Doulamis, A., Doulamis, N.: Interpretable deep learning framework for land use and land cover classification in remote sensing using shap. IEEE Geoscience and Remote Sensing Letters  \textbf{20}, ~1--5 (2023)

\bibitem{hyperspectral_vegetation}
Thenkabail, P.S., Enclona, E.A., Ashton, M.S., Van Der~Meer, B.: Accuracy assessments of hyperspectral waveband performance for vegetation analysis applications. Remote Sens. Environ.  \textbf{91}(3-4),  354--376 (Jun 2004)

\bibitem{sentinel1}
Torres, R., Snoeij, P., Geudtner, D., Bibby, D., Davidson, M., Attema, E., Potin, P., Rommen, B., Floury, N., Brown, M., et~al.: Gmes sentinel-1 mission. Remote sensing of environment  \textbf{120},  9--24 (2012)

\bibitem{unesco2021recommendation}
UNESCO, C.: Recommendation on the ethics of artificial intelligence (2021)

\bibitem{sa_crop_type_dataset}
{Western Cape Department of Agriculture}, {Radiant Earth Foundation}: Crop type classification dataset for western cape, south africa (2021)

\bibitem{Xie_2015_ICCV}
Xie, S., Tu, Z.: Holistically-nested edge detection. In: Proceedings of the IEEE International Conference on Computer Vision (ICCV) (December 2015)

\bibitem{time_series_crop_classification}
Zhong, L., Hu, L., Zhou, H.: Deep learning based multi-temporal crop classification. Remote Sensing of Environment  \textbf{221},  430–443 (Feb 2019). \doi{10.1016/j.rse.2018.11.032}, \url{http://dx.doi.org/10.1016/j.rse.2018.11.032}

\end{thebibliography}
